\documentclass[nohyperref]{article}

\usepackage{microtype}
\usepackage{graphicx}
\usepackage{subfigure}
\usepackage{booktabs} 

\usepackage{hyperref}
\usepackage{times}
\usepackage{latexsym}
\usepackage{amssymb}
\usepackage{amsfonts}
\usepackage{amsmath} 
\usepackage{amsthm} 
\usepackage{booktabs}
\usepackage{enumerate}
\usepackage{enumitem}
\usepackage{graphicx}
\usepackage{subfigure}
\usepackage{xspace}
\usepackage{float}
\usepackage{bbm}
\usepackage{bm}
\usepackage{multirow}
\usepackage{booktabs}
\usepackage{color}
\usepackage{framed}
\usepackage{stfloats}
\usepackage{iitem}
\usepackage{makecell}
\usepackage{float}
\usepackage{placeins}
\usepackage{afterpage}
\usepackage{bbding}
\usepackage[normalem]{ulem}
\usepackage{soul}
\usepackage{xcolor}
\usepackage{algorithm}
\usepackage{array}
\usepackage{pifont}
\usepackage{listings}

\usepackage{url}
\newcommand{\paratitle}[1]{\vspace{1.5ex}\noindent\textbf{#1}}
\newcommand{\ie}{\emph{i.e.,}\xspace}
\newcommand{\aka}{\emph{a.k.a.,}\xspace}
\newcommand{\eg}{\emph{e.g.,}\xspace}

\newcommand{\ignore}[1]{}
\newcommand{\tabincell}[2]{\begin{tabular}{@{}#1@{}}#2\end{tabular}}

\usepackage{xcolor}
\definecolor{tOrange}{RGB}{255,165,0}
\definecolor{tBlue}{rgb}{0.39,0.58,0.93}
\definecolor{tPink}{RGB}{255,20,147}
\definecolor{tGreen}{RGB}{50,205,50}
\definecolor{tGold}{RGB}{255,215,0}
\newcommand{\cmark}{\ding{51}}
\newcommand{\xmark}{\ding{55}}

\usepackage[accepted]{icml2022}

\usepackage{amsmath}
\usepackage{amssymb}
\usepackage{mathtools}
\usepackage{amsthm}

\usepackage[capitalize,noabbrev]{cleveref}

\theoremstyle{plain}

\theoremstyle{definition}

\theoremstyle{remark}

\usepackage[textsize=tiny]{todonotes}

\icmltitlerunning{GlyphDiffusion: Text Generation as Image Generation}

\begin{document}

\twocolumn[
\icmltitle{\emph{GlyphDiffusion}: Text Generation Is Also Image Generation}




\begin{icmlauthorlist}
\icmlauthor{Junyi Li}{gsai,udem,bdai}
\icmlauthor{Wayne Xin Zhao}{gsai,bdai}
\icmlauthor{Jian-Yun Nie}{udem}
\icmlauthor{Ji-Rong Wen}{gsai,bdai,info}
\end{icmlauthorlist}

\icmlaffiliation{gsai}{Gaoling School of Artificial Intelligence, Renmin University of China}
\icmlaffiliation{udem}{DIRO, Université de Montréal}
\icmlaffiliation{bdai}{Beijing Key Laboratory of Big Data Management and Analysis Methods}
\icmlaffiliation{info}{School of Information, Renmin University of China}

\icmlcorrespondingauthor{Wayne Xin Zhao}{batmanfly@gmail.com}

\icmlkeywords{Machine Learning, ICML}

\vskip 0.3in
]



\printAffiliationsAndNotice{\icmlEqualContribution} 

\begin{abstract}
Diffusion models have become a new generative paradigm for text generation. Considering the discrete categorical nature of text, in this paper, we propose \textsc{GlyphDiffusion}, a novel diffusion approach for text generation via text-guided image generation. Our key idea is to render the target text as a \emph{glyph image} containing visual language content. In this way, conditional text generation can be cast as a glyph image generation task, and it is then natural to apply continuous diffusion models to discrete texts. Specially, we utilize a cascaded architecture (\ie a base and a super-resolution diffusion model) to generate high-fidelity glyph images, conditioned on the input text. Furthermore, we design a text grounding module to transform and refine the visual language content from generated glyph images into the final texts. In experiments over four conditional text generation tasks and two classes of metrics (\ie quality and diversity), \textsc{GlyphDiffusion} can achieve comparable or even better results than several baselines, including pretrained language models. Our model also makes significant improvements compared to the recent diffusion model. 

\end{abstract}

\section{Introduction}
\label{sec-intro}

Diffusion models~\cite{sohl2015deep} are a class of generative models that have recently  shown to be powerful in 
synthesizing high-quality image~\cite{imagen}, audio~\cite{KongPHZC21} and video~\cite{ho2022imagen}.
They are trained to gradually transform random noise drawn from a Gaussian distribution into a sample from an unknown data distribution specified by a collection of samples.
Compared to existing generative models such as GAN~\cite{gan}, VAE~\cite{vae}, and flow-based models~\cite{flow}, diffusion models offer several desirable properties such as distribution coverage,
a stationary training objective, and easy scalability~\cite{song2019generative,dhariwal2021diffusion}. It has been shown  that diffusion models are theoretically underpinned by non-equilibrium thermodynamics and score-based generative models~\cite{ddpm,nichol2021improved}.

Although diffusion models have made great success in the vision and audio domains~\cite{KongPHZC21,imagen,dall-e}, it remains an open challenge to extend diffusion models to natural language due to the inherently discrete nature of texts. Consequently, prior work has focused on developing  approaches based on discrete diffusion by introducing transition matrices between tokens to corrupt and recover texts~\cite{austin2021structured,he2022diffusionbert,reid2022diffuser}. However, these methods cannot benefit from the improvements made on continuous diffusion models. 
Another line of work considers  continuous text representations (\eg word embedding or hidden states) as training target, and learns diffusion models in the corresponding semantic  space~\cite{li2022diffusion,gong2022diffuseq,strudel2022self,lin2022genie}.
However, unlike the target is usually fixed for continuous data (\eg image and audio), such training targets need to be learned from scratch for discrete texts, and they also correspond to different representations depending on the pre-trained models. 
Thus, it might cause the collapse of the denoising loss function and bring instability to the training process~\cite{gao2022difformer}. 

In this paper, we propose \textsc{GlyphDiffusion}, a novel text generation approach via  text-guided image generation based on continuous diffusion models. 
The key idea is that we render a target text as an image containing visual language content (called \emph{glyph image}). In this way, the conditional text generation task can be cast as a glyph image generation task, where the glyph image is expected to contain the generated text content in a visual form conditioned on the input. This approach can naturally leverage continuous diffusion models and the fixed target (\ie glyph image) can avoid simultaneous changes in model predictions and ground truth to solve the collapse of the denoising loss.


\ignore{new continuous diffusion model for text generation without converting discrete tokens into word embeddings. In our approach, the target text is rendered as an RGB image, and the text generation task is naturally transformed into an image generation task. Consequently, we can directly apply continuous diffusion models to the discrete text generation task without large modifications.}

Specially, GlyphDiffusion introduces a cascaded architecture that integrates  
a base  diffusion model and a super-resolution diffusion model for  glyph image generation. 
We conduct the  image generation based on the input semantics captured by a frozen  T5 language model~\cite{t5}. Since our goal is to produce high-quality text output that satisfies the need of the input text, we employ classifier-free guidance~\cite{ho2022classifier}
to enhance the content  fidelity of a generated glyph image. Further, to  improve the quality of the text output, we design a text grounding component to refine and transform the visual language content from generated images into the final generation results. 

\ignore{To generate the text-rendering images, GlyphDiffusion cascades a base image diffusion model and a super-resolution diffusion model for progressive generation. Motivated by the ability of previous text-to-image diffusion models to handle free-form prompts~\cite{glide,imagen}, we apply guided diffusion to our conditional text generation task. GlyphDiffusion uses a frozen T5 language model~\cite{t5} to map the input text into a sequence of embeddings. All diffusion models are conditioned on the input text
embeddings and use classifier-free guidance~\cite{ho2022classifier}, a type of guidance that interpolates between outputs from a diffusion model with and without labels. To further improve the quality of texts on generated images, we design a vocabulary grounding layer to map the words into vocabulary and revise the text. Our approach can be seen as unifying the text and image generation tasks in a different way.}

To the best of our knowledge, we are the first that adapts continuous diffusion models to text generation 
through generating glyph images. 
While conceptually and intuitively simple, our model yields surprisingly strong results. Compared to AR and NAR models, GlyphDiffusion obtains over 50\% improvements in metrics such as BLEU and ROUGE-L. Our model outperforms prior diffusion models on text generation tasks in terms of quality and diversity (\eg $+2.54$ BLEU in Quasar-T and $+2.24$ Diverse-4 in GYAFC).

\section{Background}
\label{sec-pre}

\paratitle{Diffusion Models.} Diffusion models are a class of generative models that transform Gaussian noise into samples based on the learned data distribution via an iterative denoising process~\cite{sohl2015deep,ddpm}. Given a sample from the data distribution $\bm{x}_0 \sim q(\bm{x}_0)$, the \textit{forward process} of diffusion models produces a Markov chain of latent variables $\bm{x}_1,...,\bm{x}_T$ by adding Gaussian noise to the sample:
\begin{equation}\label{eq-forward}
    q(\bm{x}_t|\bm{x}_{t-1}) = \mathcal{N}(\bm{x}_t;\sqrt{1-\beta_t}\bm{x}_{t-1},\beta_t \bm{I}),
\end{equation}
where $\beta_1,...,\beta_T$ are small enough noise levels that make $\bm{x}_T$  well approximated by $\mathcal{N}(\bm{0}, \bm{I})$. This parametrization gives us a closed form to sample any $\bm{x}_t$ given $\bm{x}_0$:
\begin{equation}
    q(\bm{x}_t|\bm{x}_0) = \mathcal{N}(\bm{x}_t;\sqrt{\bar{\alpha}_t}\bm{x}_0, 1-\bar{\alpha}_t\bm{I}), \label{eq-forward-sim}
\end{equation}
where $\alpha_t=1-\beta_t$, $\bar{\alpha}_t=\prod_{s=1}^t\alpha_s$. We can further compute the posterior $q(\bm{x}_{t-1}|\bm{x}_t,\bm{x}_0)$ using Bayes theorem:
\begin{align}
    q(\bm{x}_{t-1}|\bm{x}_t,\bm{x}_0) = \mathcal{N}(\bm{x}_{t-1};\tilde{\mu}_t(\bm{x}_t,\bm{x}_0), \tilde{\beta}_t\bm{I}), \label{eq-posterior} \\
    \tilde{\mu}_t(\bm{x}_t,\bm{x}_0) =~ \frac{\sqrt{\bar{\alpha}_{t-1}}\beta_t}{1-\bar{\alpha}_t}\bm{x}_0 + \frac{\sqrt{\alpha_t}(1-\bar{\alpha}_{t-1})}{1-\bar{\alpha}_t}\bm{x}_t. \nonumber \label{eq-mu}
\end{align}

For generation, the diffusion model is trained to reverse this forward process. The \textit{reverse process} starts from a Gaussian noise $\bm{x}_T \sim \mathcal{N}(0,\bm{I})$ and gradually denoise $\bm{x}_t$ 
with learned Gaussian transition (parameterized by $\theta$):
\begin{equation}
    p_\theta(\bm{x}_{t-1}|\bm{x}_t) = \mathcal{N}(\bm{x}_{t-1};\bm{\mu}_\theta(\bm{x}_t,t),\bm{\Sigma}_\theta(\bm{x}_t,t)). \label{eq-prior}
\end{equation}
The reverse process is trained to match the joint distribution of the forward process by optimizing the variational lower bound (VLB).
The VLB objective can be estimated using the posterior $q(\bm{x}_{t-1}|\bm{x}_t,\bm{x}_0)$ in Eq.~\ref{eq-posterior} and the prior $p_\theta(\bm{x}_{t-1}|\bm{x}_t)$ in Eq.~\ref{eq-prior}. 
To parameterize $p_\theta(\bm{x}_{t-1}|\bm{x}_t)$ in reverse process, the most straight method is to predict $\bm{\mu}_\theta(\bm{x}_t,t)$ with a neural network.
However, \citet{ddpm} have shown that predicting the noise $\bm{\epsilon}$ works much better. So the final objective can be simplified using the reweighted bound as follows:
\begin{equation}\label{eq-simple}
    L_\text{simple}(\theta)= \mathbb{E}_{\bm{x}_0,\bm{\epsilon},t}(\Vert \bm{\epsilon} - \bm{\epsilon}_\theta(\bm{x}_t,t) \Vert_2^2). 
\end{equation}

This objective is equal to optimizing a reweighted VLB on the data log-likelihood and has a connection to generative score matching~\cite{song2019generative,song2020score}. 
To compute this surrogate objective, we generate samples $\bm{x}_t \sim q(\bm{x}_t|\bm{x}_0)$ by applying Gaussian noise $\bm{\epsilon}$ to $\bm{x}_0$ then train a model $\bm{\epsilon}_\theta$ to predict the added noise using Eq.~\ref{eq-simple}. 

\paratitle{Diffusion Models for Conditional Generation.} 
In conditional generation, the data $\bm{x}_0$ is associated with a condition $\bm{c}$, for example a label in the case of class-conditional generation~\cite{ho2022cascaded}, a low-resolution image for super-resolution~\cite{saharia2021image}, or a text prompt in text-guided generation~\cite{dall-e}.
The goal is to learn a conditional diffusion model $p_\theta(\bm{x}_0|\bm{c})$.
Therefore, the input condition $\bm{c}$ is included into the reverse process $p_\theta(\bm{x}_{t-1}|\bm{x}_t,\bm{c})$ for deriving a new reweighted objective:
\begin{align}
    L_\text{simple}(\theta) = \mathbb{E}_{\bm{x}_0,\bm{\epsilon},t}(\Vert \bm{\epsilon} - \bm{\epsilon}_\theta(\bm{x}_t,t,\bm{c}) \Vert_2^2). \label{eq-condition-simple}
\end{align}
During training, the data $\bm{x}_0$ and the condition $\bm{c}$ are sampled jointly from the data distribution $q(\bm{x}_0, \bm{c})$, and the forward process $q(\bm{x}_{1:T}|\bm{x}_0)$ remains unchanged. The only change required is to add the condition $\bm{c}$ as an extra input to the neural network in reverse process $p_\theta(\bm{x}_{t-1}|\bm{x}_t,\bm{c})$.

\section{\textsc{GlyphDiffusion}}

\begin{figure}[tb]
	\centering
	\includegraphics[width=0.49\textwidth]{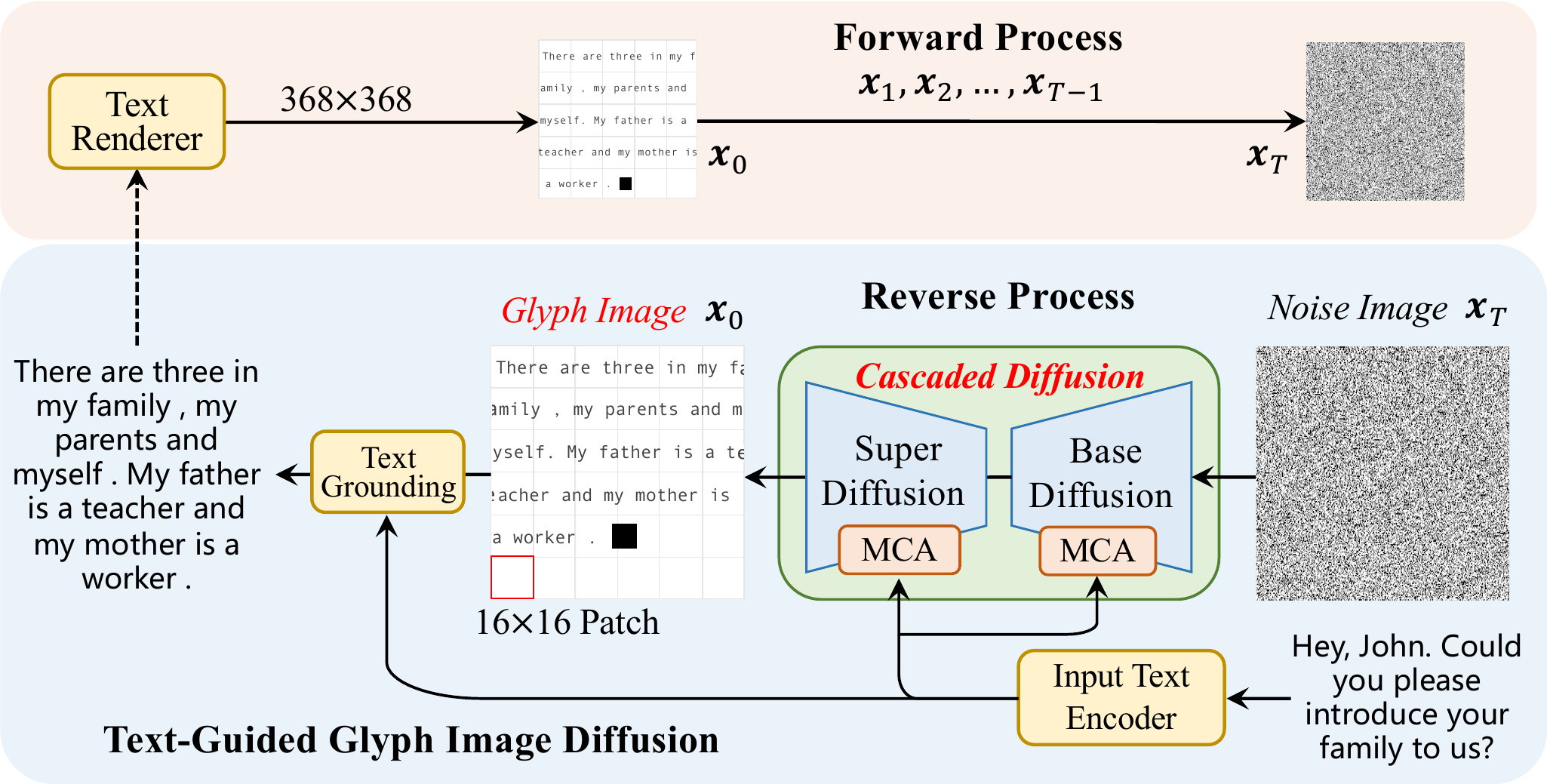}
	\caption{Overview of our proposed model \textsc{GlyphDiffusion}. ``MCA'' denotes multi-head cross attention.}
	\label{fig-model}
\end{figure}

In this section, we present \textsc{GlyphDiffusion} that casts conditional text generation as \emph{text-guided image generation}, by  establishing  the semantic map from   text condition to visual language content based on diffusion models.  The overall sketch  of our approach is shown in Figure~\ref{fig-model}.
 


\subsection{Overview}
\label{sec-renderer}


\ignore{To apply diffusion models to text generation, existing work mostly focused on 
learning to reconstruct the \emph{continuous surrogates} of a text, such as  word embeddings~\cite{li2022diffusion,gong2022diffuseq} or hidden states~\cite{lovelace2022latent}, due to the discrete nature of text. However, these surrogates themselves also need to be learned, and have varying representations in different models. 
A recent study~\cite{gao2022difformer} has shown that using learnable training targets (\eg embeddings) may cause the collapse of the denoising loss function and bring instability to the training process.
}


\ignore{To apply diffusion models to text generation, existing work mostly focused on 
learning to reconstruct the \emph{continuous surrogates}  of a text~(\ie training targets), such as  word embeddings~\cite{li2022diffusion,gong2022diffuseq} or hidden states~\cite{lovelace2022latent}, due to the discrete nature of text. While, these training targets  themselves also need to be learned, which is likely to cause the collapse of the denoising loss function~\cite{gao2022difformer}. Different from existing work, we propose a novel approach for conditional text generation based on diffusion model, by directly learning to map  \emph{a text condition} (\ie input) into \emph{an  image containing the generated text content} (\ie output).  }

To adapt diffusion models to text generation, existing studies typically  
reconstruct \emph{continuous training targets}, \eg  word embeddings~\cite{li2022diffusion,gong2022diffuseq} and hidden states~\cite{lovelace2022latent}. Since these training targets  also need to be learned beforehand, such a method is likely to cause the collapse of the denoising loss function~\cite{gao2022difformer}. Different from prior work, we  introduce a novel approach for conditional text generation based on diffusion model, by directly learning to map  \emph{a text condition} into \emph{an  image containing the generated text content}.





\paratitle{Task Formulation.} Formally, given an input text (\aka a condition) $\bm{c}$, the conditional text generation task aims to generate an output text $\bm{w}=\{w_1,...,w_n\}$ that consists of a sequence of words. However, in our approach, we consider a two-stage generation approach by incorporating an intermediate image $\bm{x}$ containing the target text $\bm{w}$: condition $\bm{c}$ $\rightarrow$ image $\bm{x}$ $\rightarrow$ text $\bm{w}$, where the two stages are implemented by a text-guided  image diffusion model $f(\cdot)$ 
and a text grounding model $g(\cdot)$, respectively. 
To discriminate the images in our setting from general images, we refer to them as \emph{glyph images}. 
Our focus lies in the first stage by training a capable glyph image diffusion model $f(\cdot)$, so as to generate high-quality language content in the visual form. 
Further, the text grounding model $g(\cdot)$ refines and transforms the visual content into the final text output $\hat{\bm{w}}$. 

\ignore{
Following previous work~\cite{gong2022diffuseq}, we focus on conditional text generation such as dialogue. Formally, given an input text (\aka a condition) $\bm{c}$, the conditional text generation task aims to generate an output text $\bm{w}=\{w_1,...,w_n\}$ that consists of a sequence of words. To apply diffusion models to text generation, in this paper, we render the output text $\bm{w}$ onto an image $\bm{x}_0 \in \mathbb{R}^{H\times W \times C}$, fundamentally distinct from transforming words into word embeddings or hidden states in previous work~\cite{li2022diffusion,gong2022diffuseq,lovelace2022latent}. Therefore, the input text $\bm{c}$ is included into the reverse process (Eq.~\ref{eq-prior}) to derive a text-guided reweighted VLB objective as follows:
}

\paratitle{Text Rendering}. To train our diffusion model, we need to prepare condition-image pairs $\langle \bm{c}, \bm{x}\rangle$ to replace condition-text pairs $\langle \bm{c}, \bm{w}\rangle$.  For this purpose, we follow \citet{rust2022language} to design a text renderer that can convert one or more pieces of text (\ie a target text in text generation datasets) into an RGB image $\bm{x} \in \mathbb{R}^{H \times W \times C}$ (taken as the \emph{target output} of diffusion models). We set the height $H=16$, the width $W=8464$, and select $C=3$ RGB input channels. In this setting, the rendered glyph image is equal to a sequence of $529$ image patches of size $16\times16$ pixels, and can be  equally converted into a square image with a $368 \times 368$ resolution (see Figure~\ref{fig-model} for an example of text rendering). 
For those texts longer than the maximum length, we truncate them as in discrete case.
 In this way, we can readily transform an existing text generation dataset to fit our setting. 



\subsection{Glyph Image Diffusion for Text Generation}

In this section, we first introduce condition encoding,  then present  text-guided glyph image diffusion, and finally describe  text grounding that maps images into text output. 

\subsubsection{Text Condition Encoding} 
\label{sec-text-encoder}

In general text-to-image diffusion models, the input texts are encoded by text encoders which can be trained on specific datasets~\cite{ramesh2021zero,glide}
or pretrained on large-scale image-text data~\cite{clip,dall-e}. 
Since they consider natural images for generation, the goal of text encoders is to encode visually meaningful and relevant semantics from input texts. 
By contrast, in our approach, the image to be generated is a rendering image only containing glyph features. Therefore, without considering visual features, we adopt pretrained text language models (\eg BERT~\cite{bert} and T5~\cite{t5})
as text encoder to capture the semantics from the condition.


Compared to image-text pre-trained models~\cite{jia2021scaling,radford2021learning}, language models are pretrained on text corpus substantially larger than paired image-text data, thus being exposed to very rich and diverse distribution of text and having a strong capability of deep textual understanding. In this paper, we explore the T5-Base model as our input text encoder, which can achieve decent performance in our experiments. We leave scaling the text encoder size for an improvement as a future work. 
Since text encoder mainly aims to inject text semantics, following previous work~\cite{imagen,dall-e}, we  freeze the parameters of text encoder during training. 


\subsubsection{Text-Guided Glyph Image Diffusion}
\label{sec-diffusion}

Since we consider glyph image specially capturing language content, it is infeasible to reuse or fine-tune prior general text-to-image models~\cite{glide,dall-e} for text generation in our approach. 
In order to generate high-fidelity images containing clear glyphs, 
we adopt a cascaded architecture~\cite{ho2022cascaded} to model the reverse process $p_\theta(\bm{x}_{t-1}|\bm{x}_t,\bm{c})$ for glyph image diffusion. 
 
\ignore{Suppose $\bm{z}_0$ is the low resolution counterpart for the high resolution image $\bm{x}_0$.
In the cascaded architecture, there are two diffusion models, \ie $p_\theta(\bm{z}_0)$ at the low resolution and $p_\theta(\bm{x}_0|\bm{z}_0)$ at the high resolution. The cascading pipeline forms a latent variable model for high resolution images, \ie $p_\theta(\bm{x}_0)=\int p_\theta(\bm{x}_0|\bm{z}_0)p_\theta(\bm{z}_0)d\bm{z}_0$. 
}

\paratitle{Cascaded Diffusion Achitecture}. We utilize a pipeline of a base $64 \times 64$ model and a super-resolution model that upsamples a $64 \times 64$ generated base image into a $368 \times 368$ image (the target image rendered by text renderer in Section~\ref{sec-renderer}). For both base and super-resolution models, we adopt the U-Net model~\cite{unet}, which is the current best architecture for image diffusion models, but change the attention layers to use multi-head attention~\cite{transformer}. 
To adapt U-Net to text-guided glyph image diffusion, we take input text embeddings (encoded by the text condition encoder in Section~\ref{sec-text-encoder}) as input. Each step of the U-Net network can attend to the sequence of word emeddings via multi-head cross-attention. Specifically, the condition encoder $\tau_\theta$ projects the input text $\bm{c}$ to a sequence of embeddings $\tau_\theta(\bm{c}) \in \mathbb{R}^{m \times d_\tau}$, where $m$ is the number of tokens and $d_\tau$ is the embedding dimension. The text-conditional cross-attention layer is implemented as follows:
\begin{align}
    \text{Attention}(Q,K,V) &= \text{softmax}(\frac{QK^\top}{\sqrt{d}})V, \label{eq-attention} \\
    Q=W_Q^{(i)}\psi_i(\bm{x}_t), K&=W_K^{(i)}\tau_\theta(\bm{c}), V=W_V^{(i)}\tau_\theta(\bm{c}), \nonumber
\end{align}
where $\psi_i(\bm{x}_t)$ denotes the flatten representation of the U-Net model at the $i$-th layer, $W_Q^{(i)} \in \mathbb{R}^{d \times d_\psi}$, $W_K^{(i)} \in \mathbb{R}^{d \times d_\tau}$, and $W_V^{(i)} \in \mathbb{R}^{d \times d_\tau}$ are learnable projection matrices. 

For the super-resolution model ($64 \times 64 \rightarrow 368 \times 368$), we adopt the Efficient U-Net model from \citet{imagen} for improving the memory efficiency, inference time, and convergence speed. The improved model makes several key modifications to the original architecture, such as reversing the order of downsampling and upsampling path in order to accelerate the forward pass of the U-Net.



\paratitle{Enhancing the Text Guidance}. Unlike general image generation, we rely on the visual content of the glyph image for text generation. Thus, text semantics from the input text are particularly important to consider in our approach.  
To enhance the guidance of input condition on the output, classifier guidance is proposed by equipping diffusion models with a separate classifier~\cite{dhariwal2021diffusion}. The classifier will model the conditional probability $p_\theta(\bm{c}|\bm{x}_{t-1})$ of predicting the input condition given the output.
However, this approach strengthens the impact of input condition at the expense of  output diversity. 
Thus, we adopt \textit{classifier-free guidance}~\cite{ho2022classifier} by jointly training a single diffusion model on conditional and unconditional objectives without a separate classifier model as follows:
\begin{equation}
    \bm{\hat{\epsilon}}_\theta(\bm{x}_t,\bm{c}) = w \cdot \bm{\epsilon}_\theta(\bm{x}_t,\bm{c}) + (1-w) \cdot \bm{\epsilon}_\theta(\bm{x}_t), \label{eq-classifier-free}
\end{equation}
where $\bm{\epsilon}_\theta(\bm{x}_t,\bm{c})$ is implemented by the text-guided cascaded diffusion model, $\bm{\epsilon}_\theta(\bm{x}_t)$ is realized by randomly dropping $\bm{c}$ from the diffusion model with a fixed probability (\eg 10\%), and $w \geq 1$ is the guidance weight.
By using classifier-free guidance, the objective in Eq.~\ref{eq-condition-simple} can be modified and adapt to our text-guided glyph image diffusion as:
\begin{align}
    L_\text{simple} = \mathbb{E}_{\bm{x}_0,\bm{\epsilon},t}(\Vert \bm{\epsilon} - \bm{\hat{\epsilon}}_\theta(\bm{x}_t,\bm{c}) \Vert_2^2). \label{eq-glyph-simple}
\end{align}

\subsubsection{Output Text Grounding}
\label{sec-grounding}

Once a glyph image is generated under the guidance of the text condition, we  consider transforming it into an output text.  A simple way is to employ off-the-shelf toolkits  such as optical character recognition for recognizing the words on the glyph image. However, such a way only focuses on word-level recognition and lacks an overall consideration of the text semantics, also suffering from potential issues such as incorrect word spelling. Therefore, we design a specific  text grounding model for improving the generated text.  

\ignore{The images generated by our diffusion model can be directly recognized as the final output texts via some off-the-shelf techniques such as optical character recognition. However, we propose to enhance the process of mapping images into texts with the input text condition. Therefore, we design a \emph{vocabulary grounding layer} that takes as input the generated image from diffusion models and outputs the final text. 
The advantages are twofold. 
First, the generated images can be regarded as a kind of visual templates, based on which our model can output better texts given the input condition text. Second, generating real words in vocabulary instead of directly recognizing as characters can some potential issues such as incorrect word spelling in generated images.
}

The text grounding model has a similar architecture to Transformer model~\cite{transformer}, while making special extensions that take a glyph image as input and condition on the input text.  
Specifically, it consists of three sub-layers, including multi-head self-attention (MHA), cross-attention (MCA), and feed-forward network (FFN). To feed the image as input, we flatten it into a sequence of $16 \times 16$ patches and map them to patch embeddings with dimension $D$:
\begin{equation}
    \bm{h}_0 = [\bm{x}_p^1\mathbf{E},..., \bm{x}_p^j\mathbf{E}, ..., \bm{x}_p^N\mathbf{E}] + \mathbf{E}_{pos},
\end{equation}
where $\mathbf{E} \in \mathbb{R}^{(P^2 \cdot C) \times D}$ is a learnable matricx that projects each 2D patch $\bm{x}_p^j$ into a patch embedding, $\mathbf{E}_{pos} \in \mathbb{R}^{N \times D}$ is the position embeddings, and $N$ is the number of patches described in Section~\ref{sec-renderer}. The MHA and MCA layers use the same attention layer in Eq.~\ref{eq-attention}, but  
 we apply layer normalization (LN) before each sub-layer and residual connections after each sub-layer:
\begin{align}
    \bm{\tilde{h}}_l &= \text{MHA}(\bm{h}_{l-1}, \bm{h}_{l-1}, \bm{h}_{l-1}), \label{eq-mha} \\
    \bm{\hat{h}}_l &= \text{MCA}(\bm{\tilde{h}}_l, \tau_\theta(\bm{c}), \tau_\theta(\bm{c})), \label{eq-mca}
\end{align}
where $\tau_\theta(\bm{c})$ is the text embeddings encoded by the input text encoder. The final FFN layer contains two linear layers with a GELU activation and outputs a hidden state $\bm{h}_l$.
The output of the last layer $\bm{h}_L$ will be used to compute the word probability distribution over the vocabulary as follows:
\begin{align}
    \text{Pr}(w_i|\bm{x}_0,\bm{c}) = \text{soft}&\text{max}(\bm{W}_v\bm{h}_{L}+\bm{b}_v). \label{eq-predict-word}
\end{align}
The text grounding model is trained to minimize the negative log-likelihood (NLL) loss as follows:
\begin{equation}
    L_\text{nll} = -\sum_{i=1}^n \log \text{Pr}(w_i|\bm{x}_0,\bm{c}). \label{eq-nll}
\end{equation}
Note that, during optimization, we can separately train the diffusion model and the text grounding model, enabling both components to focus on fulfilling specific goals. Here, we design a lightweight architecture for text grounding, which only consists of two layers, introducing almost negligible parameters compared to the total parameters of the text-guided cascaded diffusion model. 

\begin{table}[t]
	\small
	\centering
        \setlength\tabcolsep{2.8pt}
        \caption{Comparison of our work to existing diffusion models for text generation.}
	\label{tab:comparison}
	\begin{tabular}{lcccc}
		\toprule
		 \textbf{Models} & \textbf{\makecell[c]{Text\\Condition}} & \textbf{\makecell[c]{Learning\\Space}} & \textbf{\makecell[c]{Learning\\Target}} & \textbf{\makecell[c]{Target\\Fixed}} \\
		\midrule
            D3PM & \xmark & \multirow{3}{*}{discrete} & \multirow{3}{*}{words} & \cmark \\
            DiffusER & \cmark &  &  & \cmark  \\
            DiffusionBERT & \xmark & & & \cmark \\
            \cmidrule{1-5}
            LD4LG  & \cmark  & \multirow{4}{*}{continuous} & hidden states & \cmark   \\
            \cline{4-4}
		DiffusionLM  & \xmark  & & \multirow{3}{*}{\makecell[c]{word\\embeddings}} & \xmark \\    
            SeqDiffuSeq & \cmark & & & \xmark  \\
            DiffuSeq  & \cmark  &  & & \xmark   \\
            \cmidrule{1-5}
		GlyphDiffusion  & \cmark & continuous & images & \cmark      \\
		\bottomrule
	\end{tabular}
\end{table}

\subsection{Discussion and Learning}

\paratitle{Comparison}. Existing diffusion models for text generation can be categorized into two classes based on the modeling space. The first line of research, such as D3PM~\cite{austin2021structured}, DiffusER~\cite{reid2022diffuser}, and DiffusionBERT~\cite{he2022diffusionbert}, proposed to model the transition between words considering the discrete categories of texts. However, these models depart from the diffusion modeling framework and lose some capabilities of diffusion models designed for continuous representations. Another line of research, such as LD4LG~\cite{lovelace2022latent}, DiffusionLM~\cite{li2022diffusion}, and DiffuSeq~\cite{gong2022diffuseq}, focused on mapping words to continuous representations (\eg word embeddings), which need to be learned beforehand. Such a way is prone to the collapse of the denoising process and training instability. Our model is the first to map texts into glyph images, in which conditional text generation is cast as a glyph image generation task. We present a detailed comparison in Table~\ref{tab:comparison}.

\paratitle{Optimization}. The training procedure of GlyphDiffusion can be described as follows: given a training pair $(\bm{c}, \bm{x}_0)$, we first obtain a low-resolution image $\bm{z}_0$ of the glyph image $\bm{x}_0$ and map the text condition $\bm{c}$ to embeddings; then, we add Gaussian noise to $\bm{z}_0$ and $\bm{x}_0$ and obtain $\bm{z}_t$ and $\bm{x}_t$ using Eq.~\ref{eq-forward-sim}; finally, a neural network $\bm{\epsilon}_\theta$ is trained to predict the Gaussian noise based on $\bm{c}$, $\bm{z}_t$, $\bm{x}_t$, and time step $t$ with classifier-free guidance (Eq.~\ref{eq-classifier-free}). The diffusion model is optimized using $L_\text{simple}$ in Eq.~\ref{eq-glyph-simple}. Besides, we train the text grounding model given a training paier $(\bm{c}, \bm{x}_0, \bm{w})$, where $\bm{w}$ is the corresponding text of $\bm{x}_0$, using $L_\text{nll}$ in Eq.~\ref{eq-nll}. Algorithm~\ref{alg-training} presents the training procedure for our diffusion model. At inference time, based on the text condition, GlyphDiffusion first iteratively denoises the Gaussian noise to low-resolution glyph images, upon which the final glyph images can be generated in the same way.

\section{Experiments}

In this section, we detail the experimental setup and then highlight the main conclusions of our results.

\subsection{Experimental Setup}

\paratitle{Tasks and Datasets.} We evaluate \textsc{GlyphDiffusion} on four kinds of conditional text generation tasks and datasets. \emph{Open-domain dialogue} requires models to generate a fluent, engaging, and meaningful natural language response given previous dialogue turns between itself and one or more other participants~\cite{huang2020challenges}. We adopt the widely-used \textbf{DailyDialogue} dataset~\cite{li2017dailydialog}, which contains $13,118$ multi-turn dialogues extracted from various websites covering a wide range of daily topics. \emph{Question generation} aims to generate natural language questions which can be answered by the given contents~\cite{duan2017question}. We use the \textbf{Quasar-T} dataset~\cite{dhingra2017quasar}, consisting of $43,013$ open-domain trivia questions and their answers obtained from various internet sources. \emph{Style transfer} aims to change the stylistic manner of a text while preserving its meaning~\cite{toshevska2021review}. We test on a large dataset Grammarly's Yahoo Answers Formality Corpus (\textbf{GYAFC})~\cite{rao2018dear}, containing a total of $110$K informal/formal sentence pairs. We choose two sub-domains Entertainment\&Music and Family\&Relationship from this dataset. \emph{Paraphrase generation} involves rewriting a sentence with the same semantic meaning but a different syntactic or lexical form~\cite{li2017paraphrase}. We adopt the widely-used dataset Quora Question Pairs (\textbf{QQP}) crawled from the community question answering forum Quora with $147$K positive pairs. The statistics of these datasets are shown in Appendix~\ref{app-dataset}.

\begin{table*}[t]
	\small
	\centering
        \caption{Evaluation results on four conditional text generation tasks, \ie open-domain dialogue (DailyDialogue), question generation (Quasar-T), style transfer (GYAFC), and paraphrase generation (QQP). The best results are denoted by \textbf{bold} fonts, and the best results without pretrained language models are denoted by \underline{underline} fonts. ``FT'' means fine-tuning PLMs on this dataset.}
        \label{tab:main}
	\begin{tabular}{c r c c c c c c c}
		\toprule
		\textbf{Tasks} & \textbf{Models} & \textbf{BLEU}$\uparrow$  & \textbf{ROUGE-L}$\uparrow$ & \textbf{BERTScore}$\uparrow$ & \textbf{Dist-1}$\uparrow$ & \textbf{Self-BLEU}$\downarrow$ & \textbf{Diverse-4}$\uparrow$ & \textbf{Length} \\ 
            \midrule[0.5pt]
         \multirow{8}{*}{\tabincell{c}{Open-domain\\Dialogue}} & GRU-attention & 0.0662 & 0.2137 & 0.4545 & 0.7889 & 0.8145 & 0.1540 & 10.45 \\
            & Transformer-base & 0.0704 & 0.1990 & 0.4778 & 0.8934 & 0.4003 & 0.5777 & 20.01 \\
            \cmidrule[0.5pt]{2-9}
            & GPT2-base FT & 0.0749 & 0.2176 & 0.5223 & 0.9445 & 0.0229 & 0.9654 & 20.23 \\
            & GPT2-large FT & 0.0803 & 0.2434 & 0.5189 & \textbf{0.9502} & 0.0221 & 0.9500 & 20.33 \\
            & GPVAE-T5 FT & 0.0843 & 0.2402 & 0.5089 & 0.6634 & 0.3677 & 0.5809 & 21.90 \\
            \cmidrule[0.5pt]{2-9}
            & NAR-LevT & 0.0489 & 0.1054 & 0.4634 & 0.9233 & 0.8207 & 0.1453 & 6.43 \\
            & DiffuSeq & 0.0740 & 0.2329 & 0.5794 & 0.9490 & \uline{\textbf{0.0136}} & 0.9641  & 11.84 \\
            & GlyphDiffusion & \uline{\textbf{0.0855}} & \uline{\textbf{0.2450}} & \uline{\textbf{0.5844}} & \uline{0.9500} & 0.0200 & \uline{\textbf{0.9660}} & 13.20 \\
            \midrule[0.5pt]
            \multirow{8}{*}{\tabincell{c}{Question\\Generation}} & GRU-attention & 0.0651 & 0.2617 & 0.5222 & 0.7930 & 0.9999 & 0.3178 & 10.10 \\
            & Transformer-base & 0.0364 & 0.1994 & 0.5334 & 0.8236 & 0.8767 & 0.4055 & 12.10 \\
            \cmidrule[0.5pt]{2-9}
            & GPT2-base FT & 0.0741 & 0.2714 & 0.6052 & 0.9602 & \textbf{0.1403} & \textbf{0.9216} & 10.00 \\
            & GPT2-large FT & 0.1110 & 0.3215 & 0.6346 & \textbf{0.9670} & 0.2910 & 0.8062 & 10.00 \\
            & GPVAE-T5 FT & 0.1251 & 0.3390 & 0.6308 & 0.9381 & 0.3567 & 0.7282 & 11.40 \\
            \cmidrule[0.5pt]{2-9}
            & NAR-LevT & 0.0930 & 0.2893 & 0.5491 & 0.8914 & 0.9830 & 0.4776 & 6.93 \\
            & DiffuSeq & 0.1731 & \uline{\textbf{0.3665}} & 0.6123 & 0.9056 & 0.2789 & 0.8103 & 11.50 \\
            & GlyphDiffusion & \uline{\textbf{0.1985}} & 0.3566 & \uline{\textbf{0.6530}} & \uline{0.9137} & \uline{0.2005} & \uline{0.8334} & 14.31 \\
            \midrule[0.5pt]
            \multirow{8}{*}{\tabincell{c}{Style\\Transfer}} & GRU-attention & 0.0502 & 0.2757 & 0.3145 & 0.8390 & 0.8290 & 0.3321 & 10.34 \\
            & Transformer-base & 0.0677 & 0.2860 & 0.3232 & 0.8591 & 0.7991 & 0.3550 & 13.23 \\
            \cmidrule[0.5pt]{2-9}
            & GPT2-base FT & 0.0734 & 0.2945 & 0.4360 & 0.9477 & 0.0657 & 0.9112 & 16.50 \\
            & GPT2-large FT & 0.0757 & 0.3050 & 0.4143 & 0.9545 & \textbf{0.0530} & 0.9089 & 17.45 \\
            & GPVAE-T5 FT & 0.0803 & 0.3048 & 0.4235 & \textbf{0.9567} & 0.0901 & 0.5949 & 19.80 \\
            \cmidrule[0.5pt]{2-9}
            & NAR-LevT & 0.0538 & 0.2078 & 0.3523 & 0.9037 & 0.8343 & 0.3145 & 12.20 \\
            & DiffuSeq & 0.0729 & 0.3046 & 0.4695 & 0.9440 & 0.1023 & 0.9120 & 12.35  \\
            & GlyphDiffusion & \uline{\textbf{0.0813}} & \uline{\textbf{0.3088}} & \uline{\textbf{0.4834}} & \uline{0.9510} & \uline{0.0934} & \uline{\textbf{0.9344}} & 14.30 \\
            \midrule[0.5pt]
            \multirow{8}{*}{\tabincell{c}{Paraphrase\\Generation}} & GRU-attention & 0.1894 & 0.5129 & 0.7763 & 0.9423 & 0.9958 & 0.3287 & 8.30 \\
            & Transformer-base & 0.0580 & 0.2489 & 0.5392 & 0.7889 & 0.7717 & 0.4312 & 5.52 \\
            \cmidrule[0.5pt]{2-9}
            & GPT2-base FT & 0.1980 & 0.5212 & 0.8246 & 0.9798 & 0.5480 & 0.6245 & 9.67 \\
            & GPT2-large FT & 0.2059 & 0.5415 & 0.8363 & \textbf{0.9819} & 0.7325 & 0.5020 & 9.53 \\
            & GPVAE-T5 FT & 0.2409 & 0.5886 & \textbf{0.8466} & 0.9688 & 0.5604 & 0.6169 & 9.60 \\
            \cmidrule[0.5pt]{2-9}
            & NAR-LevT & 0.2268 & 0.5795 & 0.8344 & 0.9790 & 0.9995 & 0.3329 & 8.85  \\
            & DiffuSeq & 0.2413 & 0.5880 & \uline{0.8365} & 0.9807 & 0.2732 & 0.8641 &11.20 \\
            & GlyphDiffusion & \uline{\textbf{0.2503}} & \uline{\textbf{0.5895}} & 0.8355 & \uline{0.9810} & \uline{\textbf{0.2344}} & \uline{\textbf{0.8701}} & 12.32 \\
            \bottomrule
	\end{tabular}
\end{table*}

\paratitle{Baselines}. Following \citet{gong2022diffuseq}, we compare our  \textsc{GlyphDiffusion} model to four groups of baselines:

\begin{itemize}[leftmargin=*,itemsep=-1.5pt,topsep=4pt]
    \item \textbf{GRU} with attention~\cite{ChoMBB14} and \textbf{Transformer} \cite{transformer}. These are two popular models for conditional text generation based on the encoder-decoder architecture with the (self-)attention mechanism.
    \item \textbf{GPT-2}~\cite{gpt2} and \textbf{GPVAE}~\cite{gpvae}. They are two pre-trained language models, among which GPT-2 is trained with language modeling and GPVAE augments T5~\cite{t5} with VAE.
    \item \textbf{NAR-LevT}~\cite{levt}. It is a strong iterative non-autoregressive (NAR) text generation model that adopts two operations, \ie insertion and deletion, to generate and refine sequences iteratively.
    \item \textbf{DiffuSeq}~\cite{gong2022diffuseq}. It is the recent diffusion model specially designed for conditional text generation. It uses partially noising to model the conditional probability in a single model without a separate classifier.
\end{itemize}

We implement these models following their original papers. 
Other diffusion models~\cite{lovelace2022latent,yuan2022seqdiffuseq} present similar performance to DiffuSeq, so we select DiffuSeq as a representative. The implementation details of baselines and our model are shown in Appendix~\ref{app-configuration}.

\paratitle{Evaluation Metrics}. In text generation tasks, \emph{quality} and \emph{diversity} are two key aspects for generated texts. To evaluate the quality, we adopt two automatic metrics, \ie BLEU~\cite{papineni2002bleu} and ROUGE~\cite{lin2004rouge}, which computes the overlapping $n$-grams between generated and gold texts. Since string matching based metrics can be insufficient for open-ended generation, we use BERTScore~\cite{bertscore} to assess the semantic similarity between generated and gold texts at the embedding level. As for diversity, we adopt Distinct~\cite{distinct}, which computes the number of distinct $n$-grams in generated texts, and Diverse~\cite{diversity}, which measures the ratio of distinct $n$-grams to the total number of generated words.
In addition to token-level diversity evaluation, we use self-BLEU~\cite{zhu2018texygen}, a sentence-level metric that measures the overlapping $n$-grams among the generated texts. Following \citet{gong2022diffuseq}, we generate three samples for each text condition to compute the diversity metrics.

\subsection{Main Results}
\label{sec-full}

Table~\ref{tab:main} show the results of \textsc{GlyphDiffusion} and baselines on four conditional text generation tasks.

First, compared to vanilla auto-regressive (AR) text generation models GRU and Transformer, GlyphDiffusion can achieve better results in four tasks at all quality and diversity metrics, which demonstrates the emergent capabilities of diffusion models in text generation. 
For the NAR baseline LevT, although it can outperform vanilla AR models in some cases, our GlyphDiffusion model can always obtain better performance with large margins (over 50\% improvements on BLEU in DailyDialogue and ROUGE-L in GYAFC).

Second, compared to pretrained models GPT-2 and GPVAE-T5, GlyphDiffusion can outperform the base variants for most tasks and metrics, while achieving comparable performance to the large variants. It is worth noting that the large models have much more parameters than GlyphDiffusion to ensure high-quality generation results. As for the recent diffusion model DiffuSeq, our model wins 21 out of 24 competitions (4 tasks $\times$ 6 metrics), which indicates the effectiveness of our method that casts conditional text generation as a glyph image generation task.

Finally, in terms of diversity, GlyphDiffusion can generate significantly more diverse texts compared to AR, NAR, and pre-trained models, as shown by sentence-level diversity metrics (self-BLEU and Diverse-4). As for the word-level measure Distinct-1, we can observe that GlyphDiffusion is comparable with the pretrained GPT-2 models, indicating that our model has little repetition in word-by-word generation. To compare with DiffuSeq, our GlyphDiffusion model adopts a free way of generation -- producing glyph images (contain visual language contents) then refining as final texts based on the condition. This approach can yield more diverse texts at both sentence and word levels. 

\begin{table}[t]
	\small
	\centering
	\setlength\tabcolsep{3.5pt}
        \caption{Ablation study on GYAFC dataset.}
	\label{tab:ablation}
	\begin{tabular}{lcccc}
		\toprule
		 \textbf{Models} & \textbf{BLEU} & \textbf{BERTScore} & \textbf{Dist-1} & \textbf{Diverse-4}  \\
		\midrule
		GlyphDiffusion    & 0.0813 &  0.4834  &  0.9510 & 0.9344 \\
		\midrule
            w/o Cascaded & 0.0601 & 0.4438 & 0.9112 & 0.9011 \\
		w/o Guidance  & 0.0790 & 0.4730 & 0.9410 & 0.9219  \\
            w/o Grounding & 0.0643  & 0.4566 & 0.9220 & 0.9090 \\
            \bottomrule
	\end{tabular}
\end{table}

\subsection{Detailed Analysis} 

In this part, we conduct a series of in-depth analysis to study the effectiveness of GlyphDiffusion.



\paratitle{Ablation Study}. In Section~\ref{sec-diffusion}, we design a cascaded diffusion architecture to generate high-fidelity glyph images, and utilize the classifier-free guidance technique to enhance the text guidance. To examine their importance, we design two variants of our model: (1) \textit{w/o Cascaded} removes the super-resolution model and uses the base diffusion model to generate glyph images with a 368$\times$368 resolution; (2) \textit{w/o Guidance} removes the unconditional objective $\bm{\epsilon}_\theta(\bm{x}_t)$ from Eq.~\ref{eq-classifier-free}. Furthermore, in Section~\ref{sec-grounding}, we designed a text grounding model to improve the generated text considering the overall semantics. 
To confirm its effectiveness, we design a counterpart: (3) \textit{w/o Grounding} removes the text grounding model and directly recognize the content in glyph images as final output. The ablation results are shown in Table~\ref{tab:ablation}. We can observe that removing the cascaded pipeline suffers from a large performance drop in terms of both quality and diversity metrics. This demonstrates the effectiveness of the cascaded framework in generating high-fidelity glyph images. In addition, removing classifier-free guidance or the text grounding model results in a decreased performance, but the latter is more important. The reason might be that it may circumvent some potential issues (\eg incorrect word spelling) in glyph images and improve texts.

\begin{figure}[t]
	\centering
	\subfigure[]{
		\centering
		\includegraphics[width=0.225\textwidth]{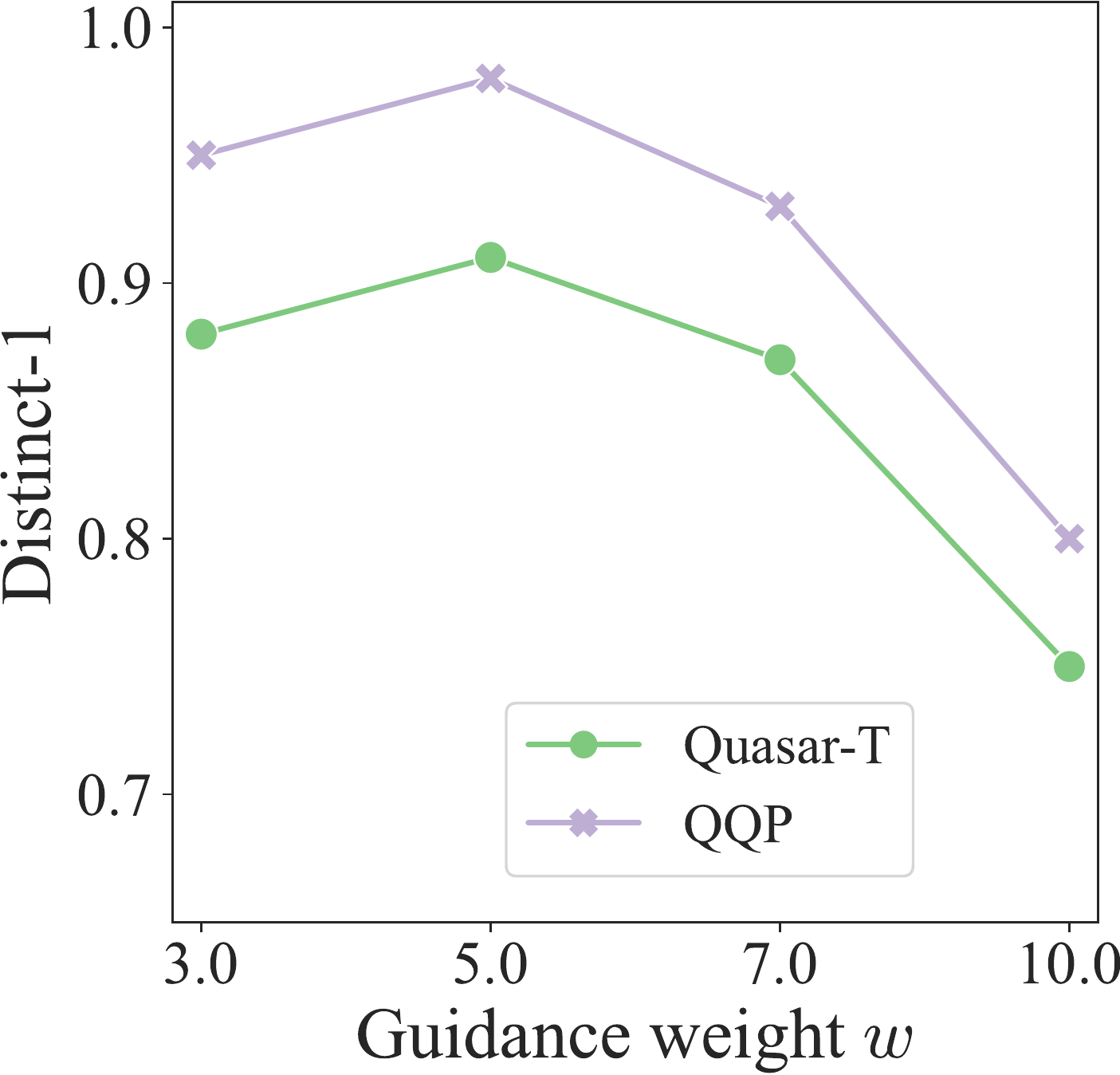}
	}
	\subfigure[]{
		\centering
		\includegraphics[width=0.22\textwidth]{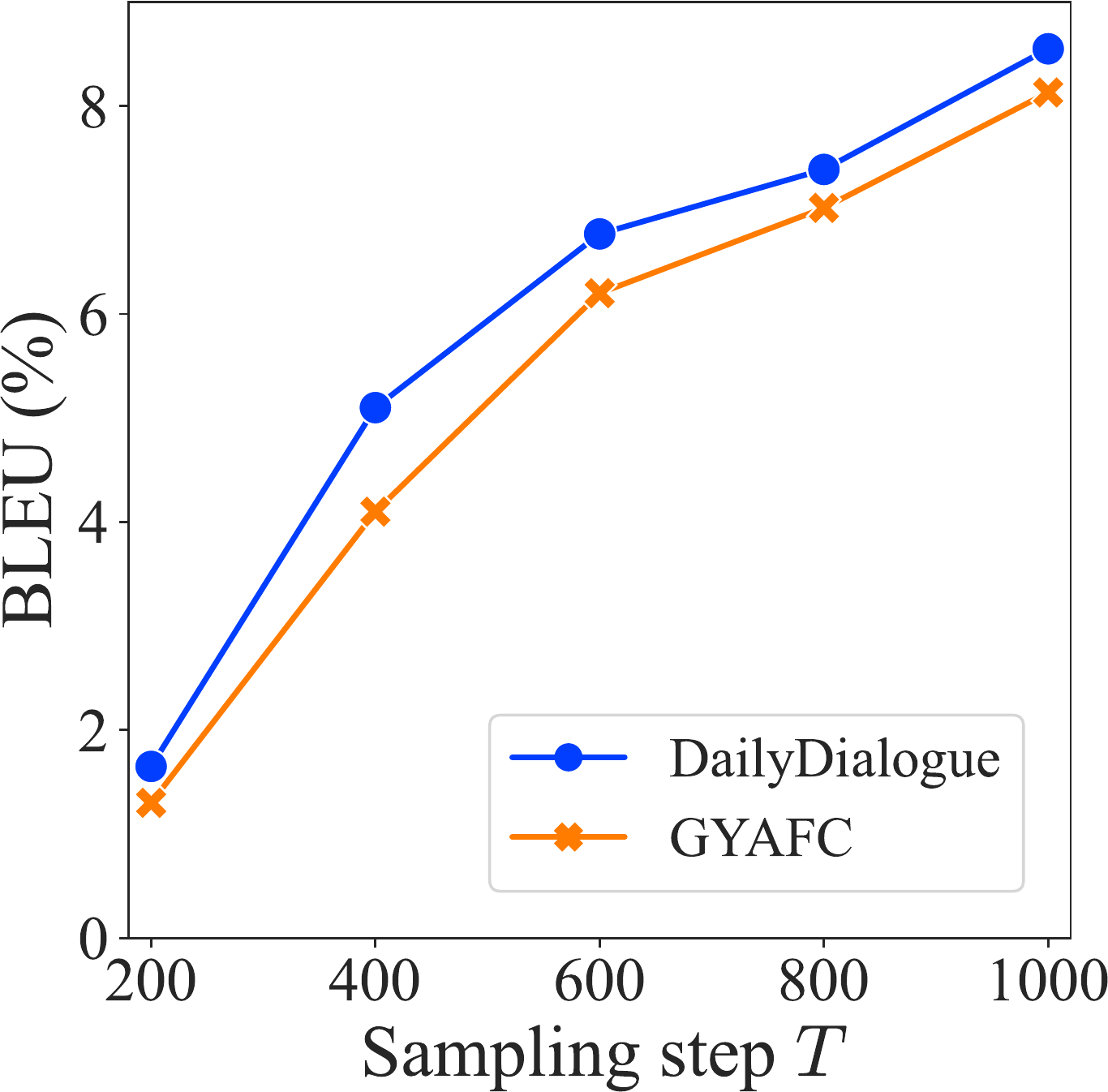}
	}
	\centering
	\caption{The Distinct-1 and BLEU scores \textit{w.r.t.} different guidance weights $w$ (a) and sampling steps $T$ (b).}
	\label{fig:sensitivity}
	\vspace{-0.2cm}
\end{figure}

\paratitle{Sensitivity Analysis}. In classifier-free guidance (Eq.~\ref{eq-classifier-free}), the weight $w$ is an important factor affecting the guidance from the text condition. A large guidance weight can improve the image-text alignment but damage the output diversity. Here, we further examine the model performance (\ie Distinct-$1$) on Quasar-T and QQP datasets by varying the guidance weight in the set \{$3.0, 5.0, 7.0, 10.0$\}.
As we can see from Figure~\ref{fig:sensitivity}(a), $w=5.0$ gives the best Distinct-$1$ score, which is the final setting in our model. While generating using larger weights (\eg $10.0$) can enhance the guidance of the condition by the super-resolution model, it gives considerably worse Distinct-$1$ (\eg $0.75$ in Quasar-T). 
The sampling step $T$ is another critical factor that significantly affects the model performance and generation speed.
Here, we fix the number of diffusion steps during training while shrinking the inference steps from $1000$ to $200$ on DailyDialogue and GYAFC datasets. 
As we can see from Figure~\ref{fig:sensitivity}(b), with the sampling step decreasing, the generated results also drop significantly (\eg from $8.55$ to $1.65$ BLEU (\%) in DailyDialogue). In practice, there is a trade-off between generation quality and inference speed.

\subsection{Case Study} 

In this section, we perform qualitative analysis to show the effectiveness of our model. In Table~\ref{tab:case-study}, we present two examples for DailyDialogue and GYAFC datasets, and the generated outputs from three baselines (\ie GPT2-base, NAR-LevT, and DiffuSeq) and our GlyphDiffusion model. As can be seen from Table~\ref{tab:case-study}, compared to NAR-LevT, our model can generate more informative and diverse texts. Since NAR-LevT adopt an iterative generation strategy, it tends to generate safe and short sentences such as ``that's all right'' in the dialogue task. As for GPT2-base which uses the powerful pretraining-finetuning paradigm, it can generate more fluent and richer content but sometimes going outside the topic of input texts. DiffuSeq sometimes generate irrelevant texts (\eg ``drink my rests''). Since we adopt a cascaded diffusion framework, our model can generate high-quality glyph images. The text grounding module can resolve some potential issues in glyph images such as repetition (\eg ``noooo'') and incorrect spelling (\eg ``gues''). More examples can be found in Appendix~\ref{app-examples}.

\begin{table}[t]
	\small
	\centering
        \setlength\tabcolsep{4pt}
        \caption{Two examples for DailyDialogue and GYAFC. We present the generations from three baselines and our model. ``w/o Grounding'' shows the content in glyph images (omitting blank patches).}
	\label{tab:case-study}
	\begin{tabular}{r|l}
		\toprule
		 \multicolumn{2}{l}{\makecell[l]{\textbf{Input: } good evening, saliva. what's that wonderful aroma \\from your kitchen? what are you doing now? [SEP] i am \\cooking now! [SEP] are you good at cooking? [SEP]}} \\
          \multicolumn{2}{l}{\makecell[l]{\textbf{Gold: }i have studied its skills recently at a training school. \\i plan to run a restaurant, so i have to practise cooking!}} \\
		 \midrule
        \textbf{GPT2-base} & \makecell[l]{no, i'm just a bad. i have a little myself, regu-\\lars, programs and more} \\
        \midrule
        \textbf{NAR-LevT} & yes. that's all right. \\
        \midrule
        \textbf{DiffuSeq}  & \makecell[l]{no, i don't drink my rests, and i need it crazy.} \\
        \midrule
        \textbf{Ours} & \makecell[l]{no, i am not good at cooking, so i need to \\practise more. it is so attractive!} \\
        \cmidrule{2-2}
        \textbf{\makecell[r]{w/o \\Grounding}} & \begin{minipage}{0.36\textwidth}\includegraphics[width=65mm, height=11mm]{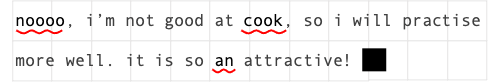}\end{minipage} \\
        \midrule\midrule
        \multicolumn{2}{l}{\makecell[l]{\textbf{Input: } its not really a book i guess but its kind a long comic.}} \\
          \multicolumn{2}{l}{\makecell[l]{\textbf{Gold: }it is a long comic, not a book.}} \\
		 \midrule
          \textbf{GPT2-base} & \makecell[l]{it's really a book, but it is seem it be a \\despite comic.} \\
        \midrule
        \textbf{NAR-LevT} & it am not really a i book females it is \\
        \midrule
        \textbf{DiffuSeq}  & \makecell[l]{not a book, but it might seem be a long comic.} \\
        \midrule
        \textbf{Ours} & \makecell[l]{not really a book, but i guess it is a long comic.} \\
        \cmidrule{2-2}
        \textbf{\makecell[r]{w/o \\Grounding}} & \begin{minipage}{0.36\textwidth}\includegraphics[width=65mm, height=5.5mm]{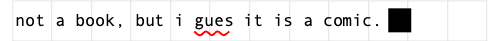}\end{minipage} \\
	   \bottomrule
	\end{tabular}
\end{table}

\section{Related Work}

\paratitle{Diffusion Models for Image Generation.} 
Diffusion models~\cite{dall-e,imagen} have demonstrated great success in generating high-quality and realistic images. Since the emergence of denoising diffusion probabilistic models (DDPM)~\cite{ddpm}, diffusion models are formalized as a forward process that corrupts the training images using Gaussian noise and a reverse denoising process that estimates the noise in the images at each step. On top of DDPM, \citet{nichol2021improved} observe that the linear noise schedule is sub-optimal for low resolution and propose a new method to avoid fast information destruction towards the end of the forward process. The work of \cite{nachmani2021non} replaces the Gaussian noise distributions with two other distributions, \ie a mixture of the Gaussian and the Gamma distribution. These works  focused on unconditional image generation without any supervision signals. By contrast, recent work has been devoted to studying text-conditioned image generation that relies on CLIP text encoding~\cite{galatolo2021generating,patashnik2021styleclip,gal2022stylegan,dall-e}. For example, \citet{abs-2110-02711} edit images with text prompts guided by a CLIP loss between the prompt and the latent. \citet{ho2022cascaded} present cascaded diffusion models, an approach for generating high-resolution images combining multiple diffusion models. Based on that, \citet{imagen} propose Imagen that uses multiple U-Net models to progressively generate high-fidelity images, which poses a similar architecture to our model. Different from prior work that generates natural images, our work renders the target texts as textual images and uses a diffusion model to generate visualized texts.
 
\paratitle{Diffusion Models for Text Generation.} To handle discrete text, prior work has extended diffusion models by defining a discrete corruption process~\cite{hoogeboom2021autoregressive,hoogeboom2021argmax}. For example, \citet{austin2021structured} and \citet{he2022diffusionbert} use transition matrices to enable gradual corruption and denoising on a sequence of discrete tokens. Unlike these works, more recent work has focused on continuous diffusion models for text~\cite{li2022diffusion,gong2022diffuseq,strudel2022self}. Diffusion-LM~\cite{li2022diffusion} works on the word embeddings and uses mapping functions to connect the
discrete and continuous space of texts. Similarly, DiffuSeq~\cite{gong2022diffuseq} is designed for sequence-to-sequence text generation using one single model to model the conditional probability. Furthermore, \citet{liu2022composable} propose a new efficient approach for composable text operations in the compact, low-dimensional latent space of text. In this paper, we also focus on continuous diffusion models for text generation but differ in that texts are rendered as continuous images instead of word embeddings. The key advantage of our method is that it allows an efficient diffusion process without a need of training an embedding step and a rounding step. Therefore, rendered text images can be an effective alternative to embeddings to leverage the continuous diffucion models. To the best of our knowledge, our work is the first to explore this setting for conditional text generation.

\section{Conclusion}

This paper presented a diffusion model, \textsc{GlyphDiffusion}, for conditional text generation. We render a target text onto a glyph image containing visual language content, so that conditional text generation can be cast as a glyph image generation task. It enables continuous diffusion models to be naturally leveraged in our approach. In order to generate high-fidelity glyph images, we introduce a cascaded diffusion architecture equipped with classifier-free guidance. Further, we design a text grounding module that can refine and transform the content from glyph images into final texts. Experiments on four conditional text generation tasks show the effectiveness of our model to previous AR, NAR, and diffusion models. In future work, we will consider applying our model to more kinds of tasks. This study proposes a new line of research using  diffusion models for text generation and demonstrates its effectiveness. Further research can explore more alternatives along the same line.

\bibliography{icml_ref}
\bibliographystyle{icml2022}

\newpage
\appendix

\section*{Appendix}
We provide some experiment-related information as supplementary materials. The appendix is organized into three sections:
\begin{itemize}
	\item Statistics of each dataset are presented in Appendix~\ref{app-dataset};
	\item Training settings of baselines and our model are presented in Appendix~\ref{app-configuration};
	\item Generated examples by our model are presented in Appendix~\ref{app-examples}.
\end{itemize}

\section{Statistics of Datasets} \label{app-dataset}
The detailed information of these four datasets is listed in Table~\ref{tab-data}.

\begin{table}[h]
	\centering
	\small
        \caption{Statistics of four datasets. \#Output denotes the average number of tokens in the output texts.}
	\label{tab-data}%
	\begin{tabular}{l|r r r r}
		\toprule[1pt]
		\textbf{Dataset} & \#Train & \#Valid & \#Test & \#Output \\
		\midrule[0.7pt]
		\textbf{DailyDialogue} & 76,052 & 7,069 & 6,740 & 13.89 \\
		\textbf{Quasar-T} & 116,953 & 2,048 & 10,000 & 10.48 \\
		\textbf{GYAFC}	 & 52,595 & 2,877 & 1,416 & 13.02 \\
            \textbf{QQP}	 & 144,715 & 2,048 & 2,500 & 9.86 \\
		\bottomrule[1pt]
	\end{tabular}%
\end{table}

\section{Implementation Details} \label{app-configuration}

\paratitle{Baseline Settings}. We follow the same baseline settings as \citet{gong2022diffuseq} and the results on Quasar-T and QQP are also collected from their work. The settings are listed in Table~\ref{tab-baseline}. For GRU-attention encoder-decoder model, we do not conduct diversity search algorithms on it, leading to poor sentence-level diversity. For NAR-LevT, we also set the max iteration to $9$ and utilize the termination condition described in the original paper. For GPVAE-T5, we set the scalars of all tasks as $2$.

\begin{table}[h]
	\centering
	\small
        \setlength\tabcolsep{3.7pt}
        \caption{The settings of different baselines. \#Para. denotes the total amount of parameters.}
	\label{tab-baseline}%
	\begin{tabular}{l|c c c}
		\toprule[1pt]
		\textbf{Models} & \textbf{\#Para.} & \textbf{\makecell[c]{Learning\\Paradigm}} & \textbf{\makecell[c]{Diversity\\Method}} \\
            \midrule[0.7pt]
            GRU & 65M & encoder-decoder & - \\
            Transformer & 80M & encoder-decoder & Temperature \\
            \midrule[0.7pt]
            GPT2-base & 117M & pretrain-finetune & Hybrid strategy \\
            GPT2-large & 774M & pretrain-finetune & Hybrid strategy \\
            GPVAE-T5 & 220M & pretrain+VAE & Gaussian sampling \\
            \midrule[0.7pt]
            NAR-LevT & 80M &  non-autoregressive & - \\
            DiffuSeq & 91M &  non-autoregressive & Gaussian sampling \\
		\bottomrule[1pt]
	\end{tabular}%
\end{table}

\paratitle{\textsc{GlyphDiffusion} Settings}. For our cascaded diffusion architecture, we follow the settings as \citet{imagen}. For the $64\times64$ base model, we use the Adafactor optimizer with a learning rate of 1e-4 for training. The hyper-parameters are set as follows:
\begin{align}
&\text{``attn\_resolutions'': [32, 16, 8]} \nonumber \\
&\text{``channel\_mult'': [1, 2, 4, 8]} \nonumber \\
&\text{``dropout'': 0} \nonumber \\
&\text{``embed\_dim'': 128} \nonumber \\
&\text{``cond\_embed\_dim'': 768} \nonumber \\
&\text{``num\_res\_blocks'': 3} \nonumber \\
&\text{``text\_cross\_attn\_res'': [32, 16, 8]} \nonumber 
\end{align}
For the $64\times64 \rightarrow 368\times368$ super-resolution model, we use an Efficient U-Net architecture for this model. Besides, we use the Adam optimizer with a learning rate of 1e-4 for training. The hyper-parameters are set as follows:
\begin{align}
&\text{``channel\_mult'': [1, 2, 4, 8]} \nonumber \\
&\text{``embed\_dim'': 128} \nonumber \\
&\text{``cond\_embed\_dim'': 768} \nonumber \\
&\text{``num\_res\_blocks'': [2, 4, 8, 8]} \nonumber  
\end{align}
For the text grounding model, we use the Adam optimizer with a learning rate of 1e-3 for training. The hyper-parameters are set as follows:
\begin{align}
&\text{``dropout'': 0.3} \nonumber \\
&\text{``embed\_dim'': 768} \nonumber \\
&\text{``ffn\_dim'': 3072} \nonumber \\
&\text{``num\_layer'': 2} \nonumber \\
&\text{``num\_head'': 12} \nonumber 
\end{align}
We present the training procedure for our diffusion model in Algorithm~\ref{alg-training}.

\section{Case Study} \label{app-examples}

We show some qualitative examples of these four datasets in Table~\ref{tab:daily}, Table~\ref{tab:quasar}, Table~\ref{tab:gyafc}, and Table~\ref{tab:qqp}. As we can see from these tables, GlyphDiffusion tends to generate good-quality and diverse texts, but still not very fluent like pretrained models.

\begin{table*}[t]
	\small
	\centering
        \setlength\tabcolsep{4pt}
        \caption{Two examples for DailyDialogue. We present the generations from three baselines and our model.}
	\label{tab:daily}
	\begin{tabular}{r|l}
		\toprule
		 \multicolumn{2}{l}{\makecell[l]{\textbf{Input: } [CLS] listen, karen, i need your help. i don't know anyone here yet. [SEP] i'm glad to help you. what's wrong?\\~[SEP] my mother - in - law just went into the hospital in l. a. hank and i will be flying there tonight. [SEP] i'm sorry to \\hear it. what's wrong with her? [SEP] doctors aren't sure yet. but the real problem is suzy. she has a bad cold, [SEP]}} \\
          \multicolumn{2}{l}{\makecell[l]{\textbf{Gold: }yes, i'd ask jill, the girl i've had before, but i need someone overnight. maybe even for two nights.}} \\
		 \midrule
        \textbf{GPT2-base} & yes, i'd ask to her and there is girl. it's number. but i know her. she is very soon. \\
        \midrule
        \textbf{NAR-LevT} & then have some do to side from and be an air. it its three and twenty and nothing domestic have to is is be hard. \\
        \midrule
        \textbf{DiffuSeq}  & \makecell[l]{i know. i'll know her and do an park. it's number. and nothing the soon to isn't you.} \\
        \midrule
        \textbf{Ours} & \makecell[l]{yes, i'd ask to her the girl. i've had before and i need someone but. maybe she is very tonight.} \\
        \midrule\midrule
        \multicolumn{2}{l}{\makecell[l]{\textbf{Input: } [CLS] thanks for inviting me to work out with you, joan. [SEP] don't mention it, let's go in. [SEP] yeah, this \\place looks great. wow, look at her, she can certainly get down, can't she? [SEP] she sure can. are you jealous, leslie?\\~[SEP] a little, i wish i could do that. [SEP] you can! with a little practice. [SEP] look at him, he's buff. [SEP] i think \\he's hot too [SEP]}} \\
          \multicolumn{2}{l}{\makecell[l]{\textbf{Gold: }that's it. i decided to turn over a new leaf. i'm going to exercise every single day.}} \\
		 \midrule
        \textbf{GPT2-base} & that's right. i don't want to make all of right now. \\
        \midrule
        \textbf{NAR-LevT} & you of that for next use to have and of my left! \\
        \midrule
        \textbf{DiffuSeq}  & \makecell[l]{if you're right, it would be true. but i don't have to have to of my bad.} \\
        \midrule
        \textbf{Ours} & \makecell[l]{that's great. i decided to go there for that. i'm supposed to make all of my wife.} \\
	   \bottomrule
	\end{tabular}
\end{table*}

\begin{table*}[t]
	\small
	\centering
        \setlength\tabcolsep{4pt}
        \caption{Two examples for Quasar-T. We present the generations from three baselines and our model.}
	\label{tab:quasar}
	\begin{tabular}{r|l}
		\toprule
		 \multicolumn{2}{l}{\makecell[l]{\textbf{Input: } [CLS] Numerous rocks and geological features abound around the 325 million year old volcano crater known \\as Arthur 's Seat . [SEP]}} \\
          \multicolumn{2}{l}{\makecell[l]{\textbf{Gold: }Edinburgh Castle stands on Arthur 's Seat what was Arthur 's seat}} \\
		 \midrule
        \textbf{GPT2-base} & what was arthur 's seat \\
        \midrule
        \textbf{NAR-LevT} & what was castle on arthur 's seat \\
        \midrule
        \textbf{DiffuSeq}  & \makecell[l]{what was castle on arthur 's seat} \\
        \midrule
        \textbf{Ours} & \makecell[l]{what was edinburgh castle on arthur 's seat} \\
        \midrule\midrule
        \multicolumn{2}{l}{\makecell[l]{\textbf{Input: } [CLS] For his discovery of human blood groups he won the 1930 Nobel Prize in Physiology or Medicine . [SEP]}} \\
          \multicolumn{2}{l}{\makecell[l]{\textbf{Gold: }Karl Landsteiner Won The Nobel Prize For Medicine In 1930 For His Discovery Of What}} \\
		 \midrule
        \textbf{GPT2-base} & for what he won the 1930 nobel prize in physiology or medicine . \\
        \midrule
        \textbf{NAR-LevT} & why he won the the 1930 physiology prize \\
        \midrule
        \textbf{DiffuSeq}  & \makecell[l]{for what he won the 1930 nobel prize in physiology or medicine .} \\
        \midrule
        \textbf{Ours} & \makecell[l]{for what he won the 1930 nobel prize in physiology or medicine .} \\
	   \bottomrule
	\end{tabular}
\end{table*}

\begin{table*}[t]
	\small
	\centering
        \setlength\tabcolsep{4pt}
        \caption{Two examples for GYAFC. We present the generations from three baselines and our model.}
	\label{tab:gyafc}
	\begin{tabular}{r|l}
		\toprule
		 \multicolumn{2}{l}{\makecell[l]{\textbf{Input: } [CLS] why do they try to sound british? [SEP]}} \\
          \multicolumn{2}{l}{\makecell[l]{\textbf{Gold: }what is the appeal of sounding british?}} \\
		 \midrule
        \textbf{GPT2-base} & why do they try to sound british? \\
        \midrule
        \textbf{NAR-LevT} & what is sounding british \\
        \midrule
        \textbf{DiffuSeq}  & \makecell[l]{why do they try to sound british?} \\
        \midrule
        \textbf{Ours} & \makecell[l]{why do they attempt to sound british?} \\
        \midrule\midrule
        \multicolumn{2}{l}{\makecell[l]{\textbf{Input: } [CLS] do u think 3 ppl in a band is a good amount?? [SEP]}} \\
          \multicolumn{2}{l}{\makecell[l]{\textbf{Gold: }do you think that having three people in a band is a good amount?}} \\
		 \midrule
        \textbf{GPT2-base} & do you think that three people location in a band is of amount? \\
        \midrule
        \textbf{NAR-LevT} & do that you think 3 peoplel in a band is a amount? \\
        \midrule
        \textbf{DiffuSeq}  & \makecell[l]{do you feel three members is a good number} \\
        \midrule
        \textbf{Ours} & \makecell[l]{do you think that three people stated in a band is enjoyable positive?} \\
	   \bottomrule
	\end{tabular}
\end{table*}

\begin{table*}[t]
	\small
	\centering
        \setlength\tabcolsep{4pt}
        \caption{Two examples for QQP. We present the generations from three baselines and our model.}
	\label{tab:qqp}
	\begin{tabular}{r|l}
		\toprule
		 \multicolumn{2}{l}{\makecell[l]{\textbf{Input: } [CLS] What is a good song to lyric prank your best friend? [SEP]}} \\
          \multicolumn{2}{l}{\makecell[l]{\textbf{Gold: }What are some good lyric prank songs to send your best friends?}} \\
		 \midrule
        \textbf{GPT2-base} & what songs with lyrics should you send to your best friends? \\
        \midrule
        \textbf{NAR-LevT} & what songs will you send your friends? \\
        \midrule
        \textbf{DiffuSeq}  & \makecell[l]{what is the songs you send to your best friends?} \\
        \midrule
        \textbf{Ours} & \makecell[l]{what lyrics songs you will send to your closest friends?} \\
        \midrule\midrule
        \multicolumn{2}{l}{\makecell[l]{\textbf{Input: } [CLS] What happens if dictatorship is continuing in the present days? [SEP]}} \\
          \multicolumn{2}{l}{\makecell[l]{\textbf{Gold: }What happens if a dictatorship continues in the present day?}} \\
		 \midrule
        \textbf{GPT2-base} & what would occur if a dictatorship continues in the present? \\
        \midrule
        \textbf{NAR-LevT} & what would happen now if a dictatorship continues? \\
        \midrule
        \textbf{DiffuSeq}  & \makecell[l]{what would happen now if a dictatorship continues?} \\
        \midrule
        \textbf{Ours} & \makecell[l]{what would happen if a dictatorship continues in the present?} \\
	   \bottomrule
	\end{tabular}
\end{table*}

\end{document}